# Moroccan Dialect -Darija- Open Dataset

**[1] Aissam Outchakoucht, [2]Hamza Es-Samaali**

[1] aissam.outchakoucht@gmail.com , [2] hamza.essamaali@gmail.com
* Equal co-authorship, order determined by coin flip

**Abstract:** Darija Open Dataset (DODa) is an open-source project for the Moroccan dialect. With more than 10,000 entries DODa is arguably the largest open-source collaborative project for Darija ⇆ English translation built for Natural Language Processing purposes. In fact, besides semantic categorization, DODa also adopts a syntactic one, presents words under different spellings, offers verb-to-noun and masculine-to-feminine correspondences, contains the conjugation of hundreds of verbs in different tenses, and many other subsets to help researchers better understand and study Moroccan dialect. This data paper presents a description of DODa, its features, how it was collected, as well as a first application in Image Classification using ImageNet labels translated to Darija. This collaborative project is hosted on GitHub platform under MIT's Open-Source license and aims to be a standard resource for researchers, students, and anyone who is interested in Moroccan Dialect.

## 1 Introduction

Nowadays, we are witnessing an unprecedented growth of IT products and services. Yet, in order for many of these solutions to flourish and be viable in a given society, they need to « understand » and be able to communicate to some extent using native languages. However, it turns out that step 0 in any serious engagement with Natural Language Processing (NLP) consists of translating the vocabulary to the widely used and most documented language in this field, namely English.

This paper presents a description of the Moroccan dialect -Darija- Open Dataset (DODa), the largest open-source collaborative project for Darija ⇆ English translation, as far as we are aware.

This open project aims to be a standard resource for researchers, students, and anyone who is interested in Moroccan Dialect. We hope for the contribution of Moroccan IT community in order to build a pedestal for any future application of NLP to benefit Moroccan people.

So far, DODa contains more than 10 thousand entries covering verbs, nouns, adjectives, verb-to-noun / singular-to-plural correspondences, conjugations, etc. The Dataset is also divided into specialized subcategories such as food, animals, human body, health, education, and many others as we will show in section 5.

## 2 Overview of the Moroccan dialect

Darija is the informal and widely used dialect in Morocco. It is spoken by over 32 million people worldwide [1]. Like the majority of Arabic dialects, Darija is severely under-resourced from a computational perspective (NLP, Machine translation, etc.). It also lacks a standard orthography to represent its sounds, in fact it only recently has begun appearing frequently in written form using either Arabic alphabet or a mixture of Latin alphabet and numbers.

A study [2] has shown that a significant amount (77,63%) of Darija vocabulary is borrowed from Modern Standard Arabic (MSA), 11,72% from French, and less than 1% from Spanish and Tamazight languages. Although those numbers may not be precise given the fact that a large part of the dataset the authors used is based on an MSA dictionary. Yet it is readily apparent that MSA provides the roots for the majority of Darija words.

For the sake of simplicity and consistency, DODa focuses (so far) on the standard Darija spoken in the regions with the highest population density in the country, and understood by the majority of Moroccans.

## 3 Related works

As for all other dialects, there are not so many research studies that have covered Darija. In this section we will focus on the study of those that have proposed large datasets for translation, namely MADAR [3] and MDED [4].

*Multi Arabic Dialect Applications and Resources (MADAR)* project consists of two subprojects covering 25 Arabic city dialects. The first is a parallel corpus in the travel domain, in which Morocco is presented through two cities (Fez and Rabat) and contains 14000 entries. The second is a lexicon of 3063 words. Darija expressions in MADAR are written using Arabic alphabet and are translated to MSA.

The second machine-readable large dataset for Darija that we are aware of is the *Moroccan Dialect Electronic Dictionary (MDED)*. It contains more than 12000 entries written using Arabic characters and translated to Modern Standard Arabic. The project, which was based mainly on the content of an MSA dictionary, also gathers a Darija-to-MSA corpus containing 34K sentences collected from different sources.

Other projects have proposed datasets such as the authors of [5] who developed a Moroccan Darija Wordnet [6] using a bilingual Moroccan-English dictionary, or websites [7] offering translated vocabulary of specific fields (usually food or cooking), not to mention printed dictionaries [8]. However, if we are seeking electronic projects with a relatively large number of entries, the existing datasets generally present significant limitations in terms of:

- *Open-source*: In fact, we believe that collecting datasets for languages, must naturally be tackled using an open and collaborative approach in order to come up with a rich and representative corpus.
- *NLP*: Be limited to nouns / verbs without handling interactions and dependences between words, ignoring conjugation, focusing on specific domains (food, travel, …) are serious limitations for using existing datasets to train real world Deep Learning models.
- *Dynamicity*: Usually, datasets are collected by one or a restricted number of people in a very limited period of time without any updates or calls for collaborations.
- *Scientific approach*: Many vocabulary datasets that one can find on the Internet are not documented, nor citable, which makes them difficult to use in research projects. Not to mention the difficulty to find and use those datasets.

## 4 Characteristics of DODa

In this section, we will try to explore DODa's features, what makes it special, and how it differs from the existing datasets:

- *Open*: We believe that data should be open to everyone, therefore DODa is hosted on GitHub [9] under MIT's Open-source license. Which provides the rights to use, copy, modify, distribute, sublicense, and/or sell copies.
- *Collaborative*: Darija is a rich and diverse dialect, therefore a collaborative approach is one of the best ways to cover this diversity. Hence, making this corpus open to collaboration is a guarantee against any potential representative bias.
- *Large*: DODa has already passed 10K entries. Each of which is written in several spellings so that it can be used for training and generalization. To the best of our knowledge, DODa is the largest Darija-English translation dataset built for Natural Language Processing purposes.
- *Growing*: One of the major benefits of being an open-source project, is the fact that it will never stop growing as far as there are contributors excited about it. Words, expressions, and new categories are and will be added continuously.
- *NLP oriented*: We are building this corpus, by design, with NLP-applications in mind. This is accomplished through the use of labels, categorization, different spellings, expressions, as well as relatively long sentences and paragraphs in the future.
- *English as a target Language*: English is the most used language in NLP. Consequently, linking Darija to a widely adopted and active language may facilitate the study of Darija, and attract the attention of many researchers to work on this dialect.
- *Alphabet*: We use an alphabet based on the Latin characters to ease the exploitation of our dataset by large spectrum of already existing models and frameworks. Besides, it is one of the two main used forms of writing of Darija and is used by a wide portion of Moroccans in social media (57% Arabic alphabet - 43% Latin alphabet [10]), without breaking the rule of one-letter-one-sound. Furthermore, we believe that it is much easier to develop a solution to perform Latin-to-Arabic-alphabet matching than the other way around, due to the explicit use of vowels in the former. Finally, it is worth noting that people are welcome to contribute using Arabic alphabet too.

## 5   Description of the dataset

DODa is a collaborative open dataset, at the time of writing it contains more than 10K entries, categorized according to various metrics:

5.a. *Parts of speech:* DODa have classes/files grouped based on similar grammatical properties, namely verbs, nouns, adjectives, adverbs, prepositions and determiners. Figure 1 shows the distribution of these classes within the dataset:

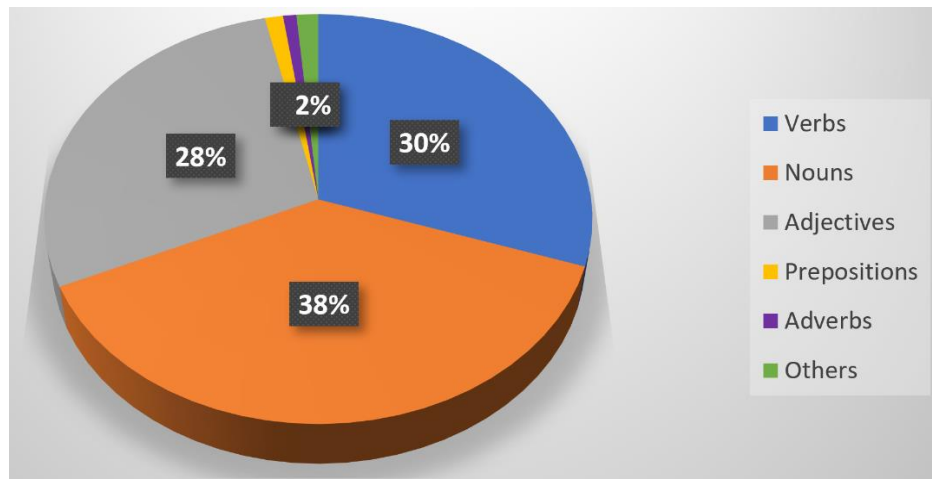

Figure 1 Parts of Speech

5.b. *Semantic:* Another type of categorization used in DODa is based on the meaning of words, therefore you'll find classes like animals, colors, food, economy, emotions, family, religion, professions, and so on. Figure 2 is an excerpt from the class *'food'*.

|    |        |        |        |        |         |
|----|--------|--------|--------|--------|---------|
| 80 | 3Tria  | 3aTria | 3Trya  | 3eTria | spices  |
| 81 | gm7    | gam7   | guem7  | 9m7    | wheat   |
| 82 | dgig   | d9i9   |        |        | flour   |
| 83 | ch3ir  |        |        |        | barley  |
| 84 | 9rfa   | 9arfa  | 9erfa  | qarfa  | cinnamon |
| 85 | kmmon  | kmmoun | kammoun | kemmon | cumin  |
| 86 | 9ronfl | 9ronfel | 9ronfal |       | cloves  |
| 87 | z3fran | z3efran | z3afran |       | saffron |
| 88 | z3tr   | ze3tr  | za3ter | z3tar  | thyme   |

Figure 2 Food class in DODa

5.c. *Beyond a lexicon:*

Apart from the classes dedicated to the translation of most used verbs, DODa also contains 3 classes that present the conjugation of more than 800 verbs translated in the past, future and imperative. The following tables give an overview of this type of files.

| 1 | root_darija | nta | nti | ntoma |
|---|---|---|---|---|
| 2 | kla | koul | kouli | koulou |
| 3 | chreb | chreb | cherbi | cherbou |
| 4 | dkhel | dkhol | dkhli | dkhlou |
| 5 | khrej | khrej | khrji | khrjou |
| 6 | fhem | fhem | fhmi | fhmou |

| 1 | ana | nta | nti | howa | hia | 7na | ntoma | homa |
|---|---|---|---|---|---|---|---|---|
| 2 | klit | kliti | kliti | kla | klat | klina | klito | klaw |
| 3 | chrbt | chrebti | chrebti | chreb | cherbat | chrebna | chrebto | cherbo |
| 4 | dkhelt | dkhelti | dkhelti | dkhel | dkhlat | dkhelna | dkhelto | dkhlo |
| 5 | khrejt | khrejti | khrejti | khrej | khrjat | khrejna | khrejto | khrjo |
| 6 | f8emt | f8emti | f8emti | f8em | f8mat | f8mna | f8mto | f8mo |
| 7 | chre7t | chre7ti | chre7ti | chre7 | chr7at | chre7na | chre7to | chr7o |

3.a. Verbs in imperative    3.b. Verbs in past tense

Figure 3 Verbs in the past and imperative within DODa

In addition, DODa contains classes matching almost 900 masculine nouns to their corresponding feminine and plural forms, as well as another class matching more than 850 verbs to their corresponding nouns as shown in Table I and II.

Table I Verb-to-noun

| Verb | Noun |
|---|---|
| **ttawa** | mtawia |
| **7TT** | 7TTan |
| **bna** | bni |
| **kla** | makla |

Table II Masculine-feminine-plural matching

| Masculine | Feminine | Masc_plural | Fem_plural |
|---|---|---|---|
| **drri** | drria | drari | drriat |
| **kas** | | kisan | |
| **ferrouj** | djaja | frarj | djajat |
| **mzyan** | mzyana | mzyanin | mzyanat |

Last but not least, the dataset also contains extra categories for idioms, proverbs and some short expressions. Figure 4 below shows the most represented classes in DODa.

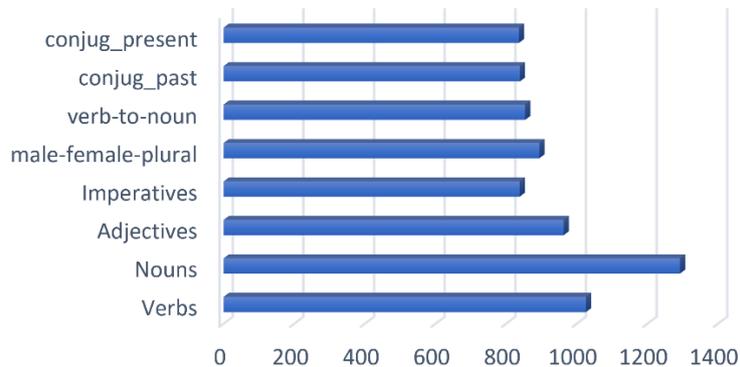

Figure 4 Most represented classes in DODa

5.d. *Guidelines:*

Of course, Darija does not have a standard orthography rules that everyone respects, yet for the sake of consistency we have tried to respect many standards that we find useful without imposing new rules that Moroccans don't already use.

    i.    In order to respect the *one-letter-one-sound* rule, we restore what Moroccans already apply as "rules" in their writing of Darija in social media (even if they are not widely used), namely numbers, capital letters, and double characters in order to write some letters that do not exist in the Latin alphabet. Table III details those correspondences.

        It should be pointed out that defining these norms explicitly aims to allow machines or models to distinguish between words that Internet users usually assume people will detect even if they are "imperfectly" written.

Table III Equivalent characters for some Arabic letters

| Darija | 3 | 7 | 9 | 8-h | 2 – 'a' – 'i' | t | T | s | S | d | D | 'ch' | 'gh' | 5 – 'kh' |
|---|---|---|---|---|---|---|---|---|---|---|---|---|---|---|
| Arabic | ع | ح | ق | ه | همزة | ت | ط | س | ص | د | ض | ش | غ | خ |

ii. Double characters to refer to the emphasis or "الشدة": (7mam – pigeons) (7mmam – bathroom)
iii. We usually don't add *"e"* in the end of Darija words: for instance, *louz* instead of *louze*
iv. We usually don't use "Z" or "*th*" for ظ , ذ , ث , because we generally don't use these letters in Darija (except in northern Morocco, but for simplicity reasons, we are focusing primarily on standard Darija)

## 6  Application: ImageNet

ImageNet [11] is a large visual database designed for use in visual object recognition. Since 2010, an annual competition, the ImageNet Large Scale Visual Recognition Challenge (ILSVRC), is organized and in which models compete to classify and detect objects and images. The challenge uses a "trimmed" list of one thousand non-overlapping classes [12].

As a first application of Darija Open Dataset project, we translated those 1000 classes to Darija using Arabic and Latin (alphanumeric) alphabet. Translations are available on DODa's repository for anyone who's aiming to implement image classification in Darija.

We actually applied a pretrained model (ResNet50 [13]) and DODa to predict the labels of random images from the Internet. Hereafter are some results:

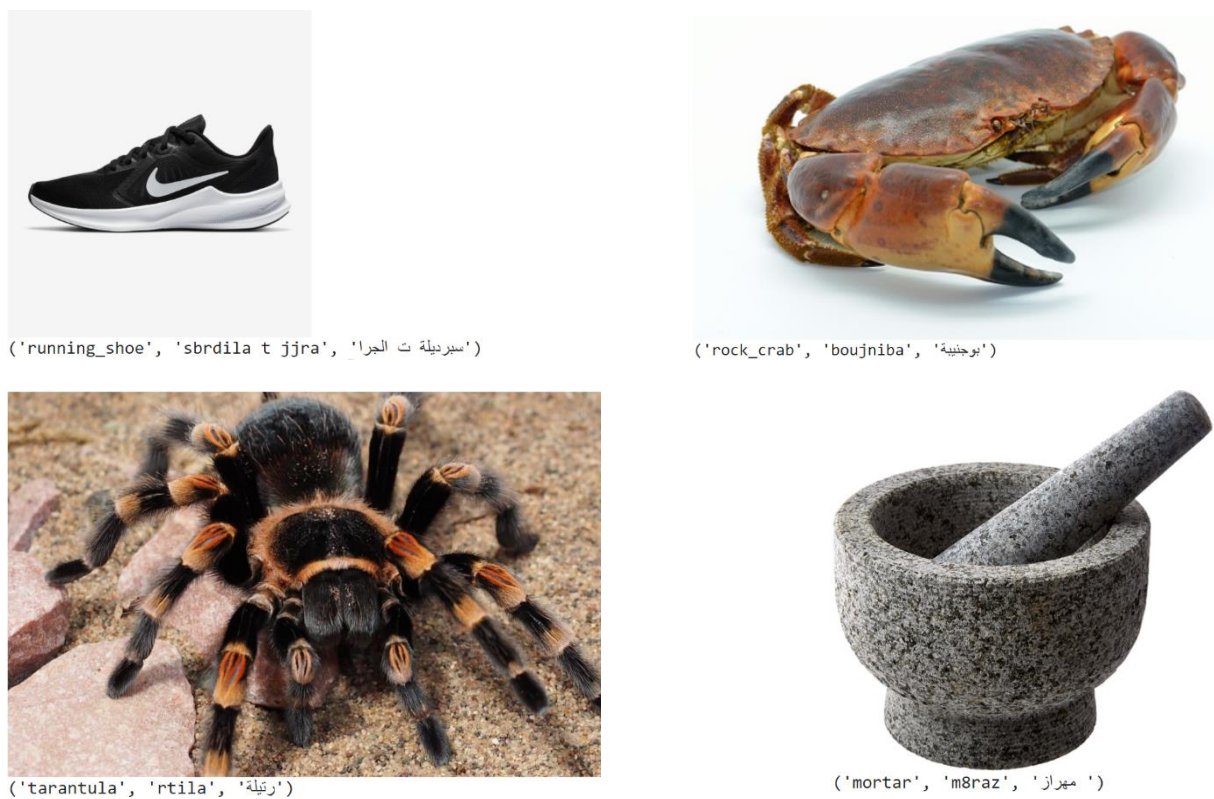

Figure 5 Random images classified using ResNet50 and DODa

## 7  Conclusion

Darija Open Dataset (DODa) is an open-source project for the Moroccan dialect. With more than 10,000 entries DODa is arguably the largest open-source collaborative project for Darija ⇆ English translation.

In this data paper, we first presented an overview of the Moroccan dialect and some related works in this field, then we detailed the characteristics and components of our dataset as well as some guidelines we respected during data collection and labelling. Finally, we presented an application of DODa in image classification by translating and using ImageNet labels and a pretrained model to classify images in Darija.

Besides being large, open and collaborative, this project is built with NLP applications in mind. To do so, we used different spellings, syntactic and semantic categorization, links and correspondences between words, conjugation, and many other concepts.

This collaborative project is hosted on GitHub platform [https://github.com/darija-open-dataset/dataset] under MIT's Open-source license and aims to be a standard resource for researchers, students, and anyone who is interested in Moroccan Dialect.